\documentclass[runningheads]{llncs}
\usepackage{graphicx}
\usepackage{amsmath,amssymb} 
\usepackage{color}

\usepackage{soul}
\usepackage{units}
\usepackage{booktabs}
\usepackage{color}
\usepackage{bm}
\usepackage{mathtools}
\usepackage{diagbox}
\usepackage{tabulary}
\usepackage{makecell}
\usepackage{verbatim} 
\usepackage{dblfloatfix}
\usepackage{array,booktabs,ragged2e}
\usepackage{subfigure}
\usepackage[percent]{overpic}
\usepackage{realboxes}
\usepackage{booktabs}

\newcommand{\citeetal}[1]{et~al.~\cite{#1}}

\def\figurePath{Figures/}
\def\myfigure#1#2{\begin{figure}[t!]\centering\includegraphics*[width = \linewidth]{\figurePath#1}\caption{#2}\label{fig:#1}\end{figure}}
\def\myfigurew#1#2#3{\begin{figure}[t!]\centering\includegraphics*[width = #3\linewidth]{\figurePath#1}\caption{#2}\label{fig:#1}\end{figure}}

\def\panelheaders#1#2#3#4{\parbox{3.0cm}{\centering\scriptsize{\fontfamily{phv}\selectfont{#1}}}\parbox{3.0cm}{\centering\scriptsize{\fontfamily{phv}\selectfont{#2}}}\parbox{3.0cm}{\centering\scriptsize{\fontfamily{phv}\selectfont{#3}}}\parbox{3.0cm}{\centering\scriptsize{\fontfamily{phv}\selectfont{#4}}}}
\def\panelfig#1{\begin{subfigure}{}\includegraphics[width=2.9cm]{\figurePath#1}\end{subfigure}}
\def\panelfiglabel#1#2{
\begin{overpic}[width=2.9cm]{\figurePath#1}
	\put(3,7){\Colorbox{black}
    {\textcolor{white}{{\centering\scriptsize{\fontfamily{phv}\selectfont{#2}}}}}}
\end{overpic}
}

\newcolumntype{R}[1]{>{\RaggedLeft\arraybackslash}p{#1}}

\newcommand{\refFig}[1]{Fig.~\ref{fig:#1}}
\newcommand{\refEq}[1]{Eq.~\ref{eq:#1}}
\newcommand{\refTbl}[1]{Tbl.~\ref{tbl:#1}}

\newcommand{\mycomment}[1]{}

\definecolor{darkred}{rgb}{0.6,0,0}
\definecolor{green}{rgb}{0.0,0.5,0}
\definecolor{blue}{rgb}{0,0,0.75}
\definecolor{orange}{rgb}{1,0.6,0.2}
\definecolor{red}{rgb}{1,0,0}

\soulregister\ref{7}
\soulregister\cite{7}
\soulregister\refFig{7}

\newcommand{\video}[1]{}

\usepackage{wrapfig}
\usepackage{hyperref}
\hypersetup{
  colorlinks, linkcolor=green
}

\begin{document}
\title{Learning to Zoom: a Saliency-Based Sampling Layer for Neural Networks} 

\titlerunning{Learning to Zoom: a Saliency-Based Sampling Layer for Neural Networks}
%
\author{Adri\`a Recasens$^*$\inst{1} \and
Petr Kellnhofer$^*$\inst{1} \and
Simon Stent\inst{2} \and
Wojciech Matusik\inst{1} \and
Antonio Torralba\inst{1}}
%
\authorrunning{A. Recasens$^*$, P. Kellnhofer$^*$ , S.Stent, W. Matusik and A.Torralba}
%

\institute{Massachusetts Institute of Technology, Cambridge MA 02139, USA 
\email{\{recasens,pkellnho,wojciech,torralba\}@csail.mit.edu}\\
Toyota Research Institute, Cambridge, MA, 02139, USA\\
\email{simon.stent@tri.global}}
\maketitle              

\begin{abstract}
We introduce a saliency-based distortion layer for convolutional neural networks that helps to improve the spatial sampling of input data for a given task.
Our differentiable layer can be added as a preprocessing block to existing task networks and trained altogether in an end-to-end fashion.
The effect of the layer is to efficiently estimate how to sample from the original data in order to boost task performance.
For example, for an image classification task in which the original data might range in size up to several megapixels, but where the desired input images to the task network are much smaller, our layer learns how best to sample from the underlying high resolution data in a manner which preserves task-relevant information better than uniform downsampling.
This has the effect of creating distorted, caricature-like intermediate images, in which idiosyncratic elements of the image that improve task performance are zoomed and exaggerated.
Unlike alternative approaches such as spatial transformer networks, our proposed layer is inspired by image saliency, computed efficiently from uniformly downsampled data, and degrades gracefully to a uniform sampling strategy under uncertainty. We apply our layer to improve existing networks for the tasks of human gaze estimation and fine-grained object classification. Code for our method is available in: \url{http://github.com/recasens/Saliency-Sampler}.
\end{abstract}

\section{Introduction}
Many modern neural network models used in computer vision have input size constraints \cite{krizhevsky2012imagenet,iandola2016squeezenet,simonyan2014very,he2016deep}. These constraints exist for various reasons. By restricting the input resolution, one can control the time and computation required during both training and testing, and benefit from efficient batch training on GPU. On certain datasets, limiting the input feature dimensionality can also empirically increase performance by improving training sample coverage over the input space.

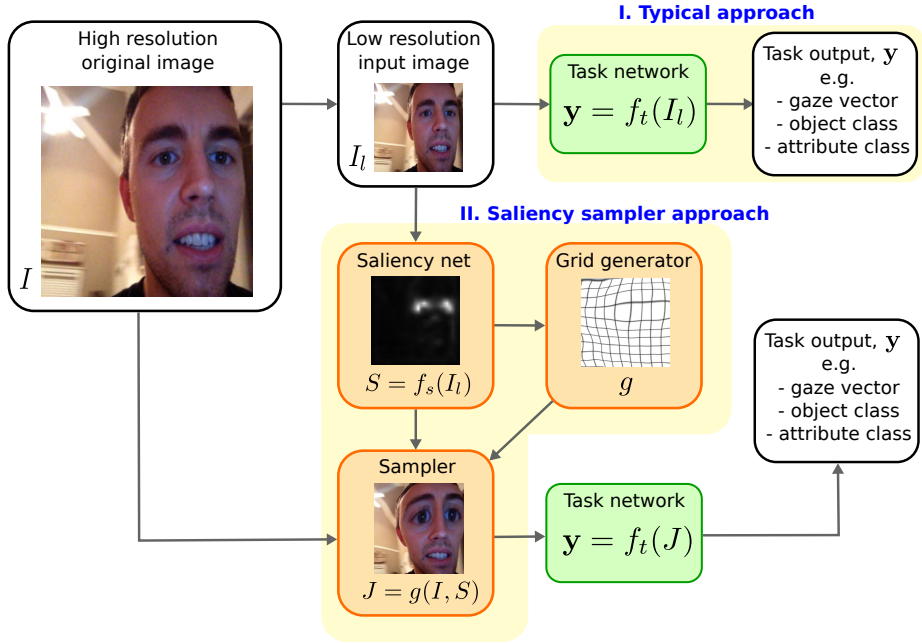
\myfigure{model_figure2}{\textbf{Outline of our proposed saliency-based sampling layer.} Numerous tasks in computer vision are solved by a task network (shown in green), operating on an image $I_l$ which has been downsampled (for performance reasons) from a much larger original image $I$. For such tasks, where $I$ is available but unused, we show that using a saliency sampler to downsample the image (rather than uniform downsampling) can lead to significant improvement in the task network performance for an identical architecture, as well as beating alternative sampling approaches such as bounding box proposals or spatial transformer networks. Our sampler is differentiable and can be trained end-to-end. The effect of the sampler is to discover and zoom in on (or sample more densely) those regions which are particularly informative to the task. In the case of gaze estimation, as seen here, the sampler locates the eyes as task-salient regions ($S$) and enlarges them in the resampled image ($J$).}

When the target input size is smaller than the images in the original dataset, the standard approach is to uniformly downsample the input images. Perhaps the best-known example is the commonly used $224\times224$ pixel input when training classifiers on the ImageNet Large Scale Visual Recognition Challenge~\cite{ILSVRC15}, despite the presence of a range of image sizes -- up to several megapixels -- within the original dataset.

While uniform downsampling is simple and effective in many situations, it can be lossy for tasks which require information from different spatial resolutions and locations.
In such cases, sampling the salient regions at the necessary (and possibly diverse) scales and locations is essential. Humans perform such tasks by saccading their gaze in order to gather the necessary information with a mixture of high-acuity foveal vision and coarser peripheral vision. Attempts have also been made to endow machines with similar forms of sampling behavior. One popular example from traditional computer vision is SIFT~\cite{lowe2004distinctive}, in which keypoints are localised within space and image scale before feature extraction. 
More recently, region proposal networks have been used widely in object detection~\cite{ren2015faster}. Mimicking the human vision system more closely, mechanisms for task-dependent sequential attention are being developed to allow numerous scene regions to be processed in high resolution (see e.g.\ \cite{itti1998model,mnih2014recurrent,ba2014multiple}). However, these approaches surrender some of the processing speed that makes machine vision attractive, and add complexity for proposal generation and evaluating task completion.

In this work we introduce a saliency-based sampling layer: a simple plug-in module that can be appended to the start of any input-constrained network and used to improve downsampling in a task-specific manner. As shown in~\refFig{model_figure2}, given a target image input size, our saliency sampler learns to allocate pixels in that target to regions in the underlying image which are found to be particularly important for the task at hand.
In doing so, the layer warps the input image, creating a deformed version in which task-relevant parts of the image are emphasized and irrelevant parts are suppressed, similar to how a caricature of a face tries to magnify those parts of a person's identity which make them stand out from the average.

Our layer consists of a saliency map estimator connected to a sampler which varies sampling density for image regions depending on their relative saliency values.
Since the layer is designed to be fully differentiable, it can be inserted before any conventional network and trained end-to-end. Unlike sequential attention models~\cite{mnih2014recurrent,ba2014multiple,eslami2016attend,fu2017look}, the computation is performed in a single pass of the saliency sampler at constant computational cost. 

\begin{figure}[t!]
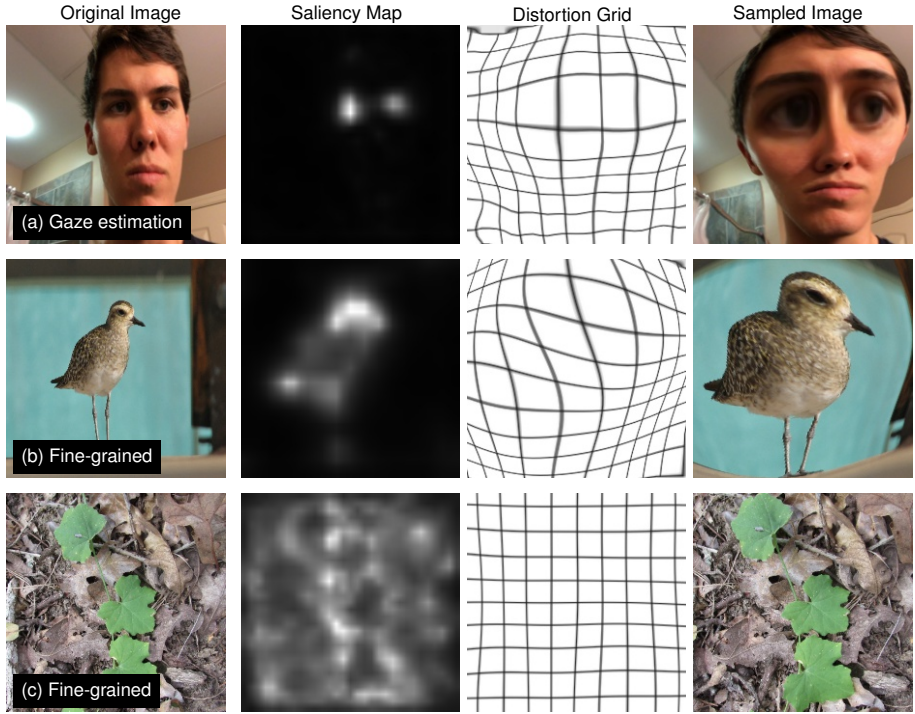

	\begin{center}
      \panelheaders{Original Image}{Saliency Map}{Distortion Grid}{Sampled Image}
		\\ 
		\panelfiglabel{013_original.jpg}{\scriptsize{(a) Gaze estimation}}
        \panelfig{013_hm.jpg}
        \panelfig{013_deformedgrid.jpg}
        \panelfig{013_deformed.jpg}
		\panelfiglabel{016_original.jpg}{\scriptsize{(b) Fine-grained}}
		\panelfig{016_hm.jpg}
		\panelfig{016_deformedgrid.jpg}
		\panelfig{016_deformed.jpg}
		\panelfiglabel{043_original.jpg}{\scriptsize{(c) Fine-grained}}
		\panelfig{043_hm.jpg}
		\panelfig{043_deformedgrid.jpg}
		\panelfig{043_deformed.jpg}
        \caption{\textbf{Examples of resampled input images for various tasks using our proposed saliency sampler.} Our module is able to discover saliency according to the task: for gaze estimation in (a), the sampler learns to zoom in on the subject's eyes to allow for higher precision gaze estimation; for fine-grained classification in (b), the sampler zooms in important parts of the bird's anatomy while cropping out much of the empty image; in (c), when no clear salient area is detected, the sampler defaults to a near-uniform sampling.}
		\label{fig:pull}
	\end{center}
\end{figure}

We apply our approach to tasks where the discovery of small objects or fine-grained details is important (see~\refFig{pull}), and consistently find that adding our layer results in performance improvements over baseline networks. 

\section{Related Work}

We divide the related work into three main categories: attention mechanisms, saliency-based methods, and adaptive image sampling methods. 

\paragraph{Attention mechanisms:}
Attention has been extensively used to improve the performance of CNNs. Jaderberg~\citeetal{jaderberg2015spatial} introduced the Spatial Transformer Network (STN), a layer that estimates a parametrized transformation from an input image in an effort to undo nuisance image variation (such as from object pose in the task of rigid object classification) and thereby improve model generalization. In their work, the authors proposed three types of transformation that could be learned: affine, projective and thin plate spline (TPS). Although our method also applies a transformation to the input image, our application is quite different: we do not attempt to undo variation such as local translation or rotation; rather we try to vary the resolution dynamically to favor regions of the input image which are more task salient. 
While our method could be encapsulated within the TPS approach of~\cite{jaderberg2015spatial}, we implicitly prevent extreme transformations and fold-overs, which can easily occur for a TPS-based spatial transformer (and which also makes direct estimation of a non-parametrized sampling map intractable). We believe that this helps to prevent dramatic failures and therefore helps to make the module easier to learn. 

Deformable convolutional networks (DCNs), introduced by Dai~\citeetal{dai2017deformable}, follow a similar motivation to STNs. They show that convolutional layers can learn to dynamically adjust their receptive fields to adapt to the input features and improve invariance to nuisance factors. Their proposal involves the replacement of any standard convolutional layer in a CNN with a deformable layer which learns to estimate offsets to the standard kernel sampling locations, conditioned on the input.
We note four main differences with our work.
First, while their method samples from the same low-resolution input as the original CNN architecture, our saliency sampler is designed to sample from any available resolution, allowing it to take advantage of higher resolution data when available. 
Second, our approach estimates the sample field through saliency maps which have been shown to emerge naturally when training fully convolutional neural networks~\cite{zhou2016learning}. We found that estimating local spatial offsets directly, as in a DCN, is much harder. 
Third, our method can be applied to existing trained networks without modification, while DCNs require changing network configurations by swapping in the deformable convolutions. 
Finally, our approach produces human readable outputs in the form of the saliency map and the deformed image which allow for easy visual inspection and debugging.
We note that our proposed saliency sampler and DCNs are not mutually exclusive: our saliency sampler is designed to sample efficiently across scale space and could potentially make use of deformable convolutional layers to help model local geometric variations.
In the same spirit as deformable networks, Li~\citeetal{li2017dense} propose an encoder-decoder structure to use non-squared convolutions. As in~\cite{jaderberg2015spatial}, they predict directly a parametrization of these transformations instead of using a saliency map. 

Attending to multiple objects recursively has also been previously explored. Eslami~\citeetal{eslami2016attend} proposed a method to iteratively attend to multiple objects in an image. In the same direction,~\cite{fu2017look} introduced a method for fine-grained classification which recursively locates an object in a low-resolution image followed by cropping from a high-resolution image. More recently,~\cite{zheng2017learning} expanded this idea to multiple attention locations in the image, instead of a single one. Finally, \cite{xiao2015application} describe a method where multiple crops are proposed and then filtered by a CNN. We note that these methods are designed specifically for classification and are not as general as our proposed sampling layer. 

\video{Cau~\citeetal{cao2017egocentric} introduce a spatio-temporal transformer module with recurrent connections between neighboring time slices for actively transforming a 3D feature map into a canonical view in both spatial and temporal dimensions.}

\paragraph{Saliency-based methods:}
CNNs have been shown to naturally direct attention to task-salient regions of the input data. Zhou~\citeetal{zhou2016learning} found that CNNs use a restricted portion of the image to inform their decision in a classification task. They proposed the use of Class Activation Maps (CAM) as a mechanism for localizing objects in images without explicit location feedback during training. 
Rosenfeld~\citeetal{rosenfeld2016visual} proposed an iterative method to crop the relevant regions of the image for fine-grained classification. They generate a CAM to highlight the regions most used by the network to make the final decision. These regions are used to crop a part of the image and generate a new CAM, which then highlights the regions of the image used by the network to inform the final prediction. As presented in \cite{zhou2016learning}, the CAM requires the use of a particular fully convolutional architecture. To overcome this limitation, \cite{selvaraju2017grad} introduces a gradient-based method to generate CAMs. Their method can be used to understand a wide variety of networks. In our work, we take advantage of the ability of CNNs to naturally localise task-salient regions, by encouraging the network to attend more to those regions.

\paragraph{Adaptive image sampling methods:}
Another possible approach to our problem is to pre-design certain feature detectors in a multi-scale strategy. This approach is usually taken when solving a particular problem where the features to use are very clear for humans. For instance, to solve the problem of gaze-tracking on a mobile device display,  Khosla~\citeetal{Khosla2016} proposed the iTracker method, a gaze estimation system based on RGB images. Their system uses the image from the device's front-facing camera, and extracts high resolution crops of both eyes and face using separate detectors. Another example along this line was presented by Wang~\citeetal{wang2017autoscaler}, who generate the features of the input image at different scales to then select the best features and produce the final output. 

Adaptive image sampling is also used for image retargeting in computer graphics \cite{rubinstein2010comparative}. Unlike in our case where the sampled image only serves as an intermediate representation for solving another problem, the goal of retargeting is to deform an image to fit a new shape and preserve content important for human observer as well as avoid visible deformations. Similarly to our concept, this can be driven by saliency \cite{wolf2007non} and formulated as an energy minimization \cite{karni2009energy} or Finite Element Method \cite{kaufmann2013finite} problem.


\section{Saliency Sampler} 
Let $I$ be a high-resolution image of an arbitrary size and let $I_l$ be a low-resolution image bounded by size $M\times N$ pixels suitable for a task network $f_t$ (\refFig{model_figure2}). Typically, CNNs rescale the input image $I$ to $I_l$ without exploiting the relative importance of $I$'s pixels. However, if our task requires information from a certain image region more than others, it may be advantageous to sample this region more densely. The saliency sampler executes this by first analyzing $I_l$ before sampling areas of $I$ proportionally to their perceived importance. In doing so, the model can capture some of the benefit of increased resolution without significant additional computational burden or risk of overfitting.

The sampling process can be divided into two stages. In the first stage, a CNN is used to produce a saliency map. This map is task specific, since different tasks may require focus on different image regions. In the second stage, the most important image regions are sampled according to the saliency map.

\subsection{Saliency Network}
The saliency network $f_s$ produces a saliency map $S$ from the low resolution image: $S = f_s(I_l)$. The choice of network for this stage is flexible and may be changed depending on the task. For all choices of $f_s$, we apply a softmax operation in the final layer to normalize the output map. 

\subsection{Sampling Approach}
Next, a sampler $g$ takes as input the saliency map $S$ along with the full resolution image $I$ and computes $J = g(I,S)$ -- that is, an image with the same dimensions as $I_l$, that has been sampled from $I$ such that highly weighted areas in $S$ are represented by a larger image extent (see \refFig{Warping}). In this section, we will discuss the possible forms that $g$ can take, and which one is more suitable for CNNs. In all cases, we compute a mapping between the sampled image and the original image and then use the grid sampler introduced in \cite{jaderberg2015spatial}. This mapping can be written in the standard form as two functions $u(x,y)$ and $v(x,y)$ such that $J(x,y) = I(u(x,y),v(x,y))$. 

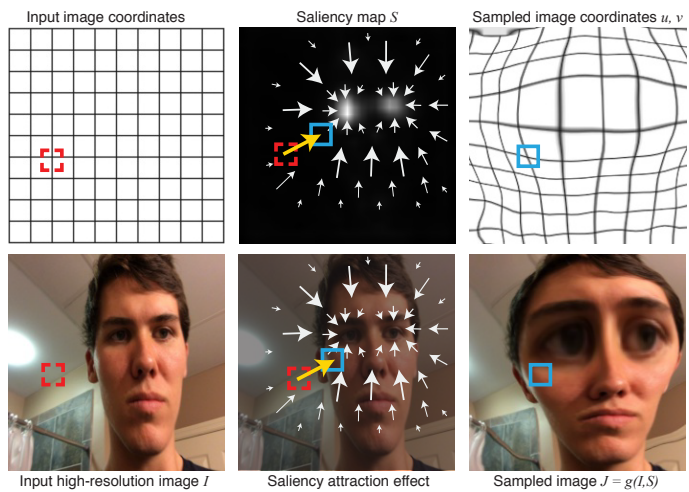
\myfigurew{Warping}{\textbf{Saliency Sampler.} The saliency map $S$ (center, top) describes saliency as a mass attracting neighboring pixels (arrows). Each pixel (red square) of the output low-resolution image $J$ samples from a location (cyan square) in the input high-resolution image $I$ which is offset by this attraction (yellow arrow) as defined by the Saliency Sampler $g(I,S)$. This distorts the coordinate system of the image and magnifies important regions which get sampled more often than others.}{0.75}

The main goal for the design of $u$ and $v$ is to map pixels proportionally to the normalized weight assigned to them by the saliency map. Assuming that $u(x,y)$, $v(x,y)$, $x$ and $y$ range from 0 to 1, an exact approximation to this problem would be to find $u$ and $v$ such that:

\begin{eqnarray}
\int_{0}^{u(x,y)} \int_{0}^{v(x,y)} S(x',y') dx' dy' = xy
\end{eqnarray}

However, finding $u$ and $v$ is equivalent to finding the change of variables that transforms the distribution set by $S(x,y)$ to a uniform distribution. This problem has been extensively explored and the usual solutions are computationally very costly~\cite{chen2010content}. For this reason, we need to take an alternative approach that is suitable for use in CNNs. 

Our approach is inspired by the idea that each pixel $(x',y')$ is pulling other pixels with a force $S(x',y')$ (see \refFig{Warping}). If we add a distance kernel $k((x,y),(x',y'))$, this can be described as:
\begin{eqnarray}
u(x,y) &=& \frac{\sum_{x',y'} S(x',y') k((x,y),(x',y')) x'}{\sum_{x',y'} S(x',y') k((x,y),(x',y'))} \label{eq:uv1} 
\\
v(x,y) &=& \frac{\sum_{x',y'} S(x',y') k((x,y),(x',y')) y'}{\sum_{x',y'} S(x',y') k((x,y),(x',y'))}
\label{eq:uv2}
\end{eqnarray}

This formulation holds certain desirable properties for our functions $u$ and $v$, notably:

\textbf{Sampled areas:} Areas of higher saliency are sampled more densely, since those pixels with higher saliency mass will attract other pixels to them. Note that kernel $k$ can act as a regularizer to avoid corner cases where all the pixels converge to the same value. In all our experiments, we use a Gaussian kernel with $\sigma$ set to one third of the width of the saliency map, which we found to work well in various settings.  

\textbf{Convolutional form:} This formulation allows us to compute $u$ and $v$ with simple convolutions, which is key for the efficiency of the full system. This layer can be easily added in a standard CNN and preserve differentiability needed for training by backpropagation. 

Note that the formulation in~\refEq{uv1} and~\refEq{uv2} has an undesirable bias to sample towards the image center. We avoid this effect by padding the saliency map with its border values.

\subsection{Training with the Saliency Sampler}

The saliency sampler can be plugged into any convolutional neural network $f_t$ where more informative subsampling of a higher resolution input is desired. Since the module is end-to-end differentiable, we can train the full pipeline with standard optimization techniques. Our complete pipeline consists of four steps (see \refFig{model_figure2}):

\begin{enumerate}
\item We obtain a low resolution version $I_l$ of the image $I$. 
\item This image is used by the saliency network $f_s$ to compute a saliency map $S = f_s(I_l)$, where task-relevant areas of the image are assigned higher weights. 
\item We use the deterministic grid sampler $g$ to sample the high resolution image $I$ according to the saliency map, obtaining the resampled image $J = g(I,S)$ which has the same resolution as $I_l$.
\item The original task network $f_t$ is used to compute our final output $y = f(J)$.
\end{enumerate}

Both $f_s$ and $f_t$ have learnable parameters and so can be trained jointly for a particular task. We found helpful to blur the resampled input image of the task network for some epochs at the beginning of the training procedure. It forces the saliency sampler to zoom deeper into the image in order to further magnify small details otherwise destroyed by the consequent blur. This is beneficial even for the final performance of the model with the blur removed.

\section{Experiments} 
In this section we apply the saliency sampler to two important problems in computer vision: gaze-tracking and fine-grained object recognition. In each case, we examine the benefit of augmenting standard methods on commonly used datasets with our sampling module. We also compare against the closest comparable methods. As an architecture for the saliency network $f_s$, in all the tasks we use ablations of ResNet-18 \cite{he2016deep} pretrained on the ImageNet Dataset \cite{deng2009imagenet} and one final $1\times1$ convolutional layer to reduce the dimensionality of the saliency map $S$. We found this network to work particularly well for classification and regression problems.

\subsection{Gaze Tracking} 
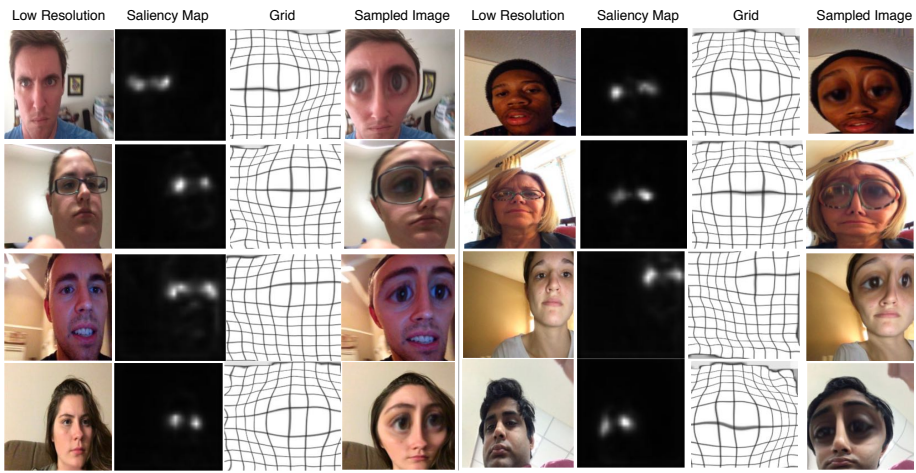
\myfigure{iTrackerResults}{\textbf{Visualization of sampler behavior for iTracker gaze-tracking task.} We show the low-resolution input image $I_l$, the saliency map $S$ estimated by $f_s$, the sampling grid $g$, and the resampled image $J$. Note that the saliency network naturally discovers the eyes to be the most informative regions in the image to infer subject gaze, but also learns to preserve the approximate position of the head in the image, which is a further useful cue for estimating gaze position on a mobile device.}

Gaze tracking systems typically focus, for obvious reasons, on the eyes. Most of the state-of-the-art methods for gaze tracking rely on eye detection, with eye patches provided to the model at the highest possible resolution. However, in this experiment we show how with fewer inputs we are able to achieve similar performance to more complex engineered systems that aim to only solve the gaze-tracking task. We benchmark our model against the iTracker dataset~\cite{Khosla2016}, and show how their original model can be simplified by using the saliency sampler. 

As the task network $f_t$ we use a standard AlexNet~\cite{krizhevsky2012imagenet} with the final layer changed to regress two outputs and a sigmoid function as the final activation. We choose AlexNet in order to be directly comparable to the iTracker system of \cite{Khosla2016}, which is one of the state-of-the-art models for gaze tracking. The model has four inputs: two crops for both eyes, one crop for the face and a coarse encoding for the location of the face in the image. As saliency network $f_S$ we use the initial 10 layers of ResNet-18. We aim to prove that our simple saliency sampler can allow a normal network to deal with the complexity of the four-input iTracker model by just magnifying the correct parts of a single input image. 

We compare our model to various competitive baselines. 
First, we replace the top three convolutional layers of an AlexNet network (pretrained in the ImageNet dataset \cite{krizhevsky2012imagenet}) with three deformable convolution layers~\cite{dai2017deformable} (\texttt{Deformable Convolutions}).
Second, we test the Spatial Transformer Network baseline~\cite{jaderberg2015spatial} with the affine parametrization (\texttt{STN}) and TPS parametrization (\texttt{STN TPS}). As a localization network, we use a network similar to $f_s$ for fairness.
Third, we modify the network $f_s$ to directly estimate the sampling grid functions $u$ and $v$ without the saliency map (\texttt{Grid Estimator}).
We also compare against the system in~\cite{Khosla2016}, which was engineered specifically for the task (\texttt{iTracker}). As an error metric, we take the average distance error of the predicted to ground truth gaze location in the screen space of the iPhone/iPad devices on which the dataset was captured. 

\begin{table}[t]
\centering
\begin{tabular}{l|c|c}
\textbf{Model} & \textbf{iPad (cm)} & \textbf{iPhone (cm)} \\ \toprule
iTracker  & 3.31 & 2.04 \\ \hline
Plain AlexNet (AN) & 5.44 & 2.63 \\ 
AN + Deformable Convolutions & 5.21 & 2.62 \\ 
AN + STN & 4.33 & 2.25\\ 
AN + STN TPS & 4.44 & 2.39 \\
AN + Grid Estimator & 3.91 & 2.20 \\ \hline
AN + Saliency Sampler (ours) & \textbf{3.29} & \textbf{2.03} \\
\end{tabular}
\caption{\textbf{Performance comparison on GazeCapture dataset.} The table reports distance errors in cm for our models and benchmarks on the GazeCapture dataset.}
\label{tab:results_itracker}
\end{table}

In Table~\ref{tab:results_itracker}, we present the performance of our model and baselines. Our model achieves a performance similar to iTracker, which enjoys the advantage of four different inputs with $224 \times 224$ pixels for each, while our system compresses all the information into a single image of $227\times 227$ pixels. Our approach also improves performance over the Deformable Convolutions, both of the STN variants and the Grid Estimator by a difference ranging from $0.62$ to $1.92$ cm for iPad and $0.17$cm to $0.59$ cm for iPhone. The STNs as well as the Deformable Convolutions have a hard time finding a transformation useful for the task, while the grid estimator is not able to find the functions $u$ and $v$ directly without the aid of a saliency map.
The intermediate outputs of our method are shown in~\refFig{iTrackerResults}. 


\subsection{Fine-Grained Classification} 
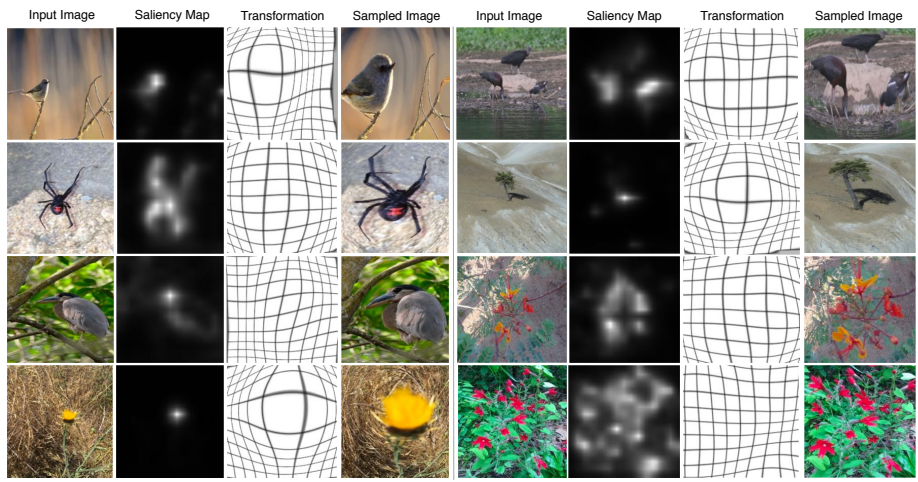
\myfigure{figure_inat}{\textbf{Visualization of sampler behavior for iNat fine-grained classification task.} Similarly to~\refFig{iTrackerResults}, the saliency network naturally discovers and zooms in on the most informative regions in the image, which tend to correspond to object parts.}

Fine-grained classification problems pose a very particular challenge: the information to distinguish between two classes is usually hidden in a very small part of the image, sometimes unresolvable at low resolution. In this scenario, the saliency sampler can play an important role: magnify the important parts of the image to preserve as many of their pixels as possible and to help the final decision network. In this experiment, we study the problem using the iNaturalist dataset that contains 5,089 animal species \cite{van2017inaturalist}. 
Our evaluation was performed using the validation set, since the test set is private and reserved for a challenge.

In this experiment, we used the ResNet-101 \cite{he2016deep} model pretrained on the ImageNet dataset \cite{deng2009imagenet} for the task network $f_t$ as it has shown a very good performance in image classification. We used an input resolution of $227\times 227$ for both the task and saliency networks, $f_t$ and $f_s$. As saliency network $f_S$ we use the initial 14 layers of ResNet-18, although the performance for other saliency networks can be found in \refTbl{new_results_inat}.

As baselines for this task, we used the same methods as before, again with ResNet-101 as the base model.
For the deformable convolutional network, we made the network modifications according to instructions in the original paper~\cite{dai2017deformable}. 
We also tested both the affine and the TPS version of STN (\texttt{STN Affine} and \texttt{STN TPS}) along with the direct grid estimator. Identically to our method, these baselines were allowed access to the original $800 \times 800$ pixel images in training time. In test time, the method had access to a center crop of $512 \times 512$ pixels. The localization networks were similar to $f_s$ for fairness.
To test whether a high-resolution input alone could improve the performance of the baseline Resnet-101 network, we also provided crops from the maximum activated regions for the ResNet-101 227 network, using the Class Activation Map method of \cite{zhou2016learning} (\texttt{CAM}). We selected the class with the largest maximal activation, and computed the bounding box as in the original paper. We then cropped this region from the original input image and rescaled it to $227\times 227$ resolution. These crops were used as inputs for the ResNet-101 $227 \times 227$ network for the final classification.

\begin{table}[t]
\centering
\begin{tabular}{l|rr|rr}
\textbf{ Model} & \textbf{Top-1(\%)} & \scriptsize{[diff]} & \textbf{Top-5(\%)} & \scriptsize{[diff]} \\ \toprule
ResNet-101 227 (RN)  			& 60 & \scriptsize{[-]} & 83 & \scriptsize{[-]}\\ \hline
RN + STN Affine                & 60 & \scriptsize{[0]} & 83 & \scriptsize{[0]}\\ 
RN + Grid Estimator           & 61 & \scriptsize{[1]} & 83 & \scriptsize{[0]}\\ 
RN + Deformable Convolutions   & 61 & \scriptsize{[1]} & 83 & \scriptsize{[0]}\\ 
RN + STN TPS                  & 62 & \scriptsize{[2]} & 84 & \scriptsize{[1]}\\ 
RN + CAM \cite{zhou2016learning}    & 62 & \scriptsize{[2]} & 84 & \scriptsize{[1]}\\ \hline
\textbf{RN + Saliency Sampler (ours)} &\textbf{65} & \textbf{\scriptsize{[5]}} & \textbf{86} & \textbf{\scriptsize{[3]}}\\ 
\end{tabular}
\caption{\textbf{iNaturalist fine-grained classification results}: top-1 and top-5 accuracy comparison on the validation set of the iNaturalist Challenge 2017 dataset.}
\label{tab:results_inat}
\end{table}

Table~\ref{tab:results_inat} shows the classification accuracy for the various models compared. 
Our model is able to significantly outperform the ResNet-101 baseline by $5\%$ and $3\%$ for top-1 and top-5 accuracies respectively. 
The performance of the CAM-based method is closer to our method which is expected as it benefits from the same idea of emphasizing image details.
However, our method still performs several points better, perhaps because of its greater flexibility to focus on local image regions non-uniformly and selectively zoom-in on certain features more than others. It also has a major benefit of being able to zoom in on an arbitrary number of non-collocated image locations, whereas doing so with crops involves determining the number of crops beforehand or having a proposal mechanism. 

The performance of the spatial transformers, the grid estimator and the deformable convolutions are similar or slightly better than the ResNet-101 baseline. Like our method, those methods benefit from the ability to focus attention on a particular region of the image. However, the affine version of the spatial transformers applies a uniform deformation across the whole image, which may not be particularly well suited to the task, while the more flexible TPS version and the grid estimator, which in theory could more closely mimic the sampling introduced by our method, were found to be harder to optimize and were consistently found to perform worse. Finally, the deformable convolutions method does not have access to the full resolution image and uses a complex parametrization which makes its training very unstable. In contrast, our method benefits from the fact that neural networks have a natural ability to predict salient image elements~\cite{zhou2015cnnlocalization} and thus the optimization may be significantly easier.

To justify our claim that the saliency sampler can benefit different task network architectures, we repeat our experiment using a Inception V3 architecture \cite{szegedy2016rethinking}. The original performance is already very high (64\% and 86\% for top-1 and top-5) as it uses higher resolution ($299$) and a deeper network, but our sampler still results in a performance of $66\%$ in top-1 and $87\%$ in top-5.

\begin{table}[h!]
\centering
\begin{tabular}{l|c|c|c|c}
 & \textbf{None (no $f_s$)} & \textbf{6-layer} & \textbf{10-layer} & \textbf{14-layer} \\ \toprule
\textbf{Top-1(\%)}  & 60 & 62 & 64 & \textbf{65} \\ 
\textbf{Top-5(\%)}  & 83 & 84 & 85 & \textbf{86} \\ 
\end{tabular}
\caption{\textbf{Saliency network ablation}: we measure the effect of different depths of saliency network $f_s$ on the iNaturalist fine-grained classification task.}
\label{tbl:new_results_inat}
\end{table}

\textbf{Saliency network importance:} In \refTbl{new_results_inat}, we retrained ResNet-101 with different depths of saliency network $f_s$. We used different ablations of ResNet-18 with 6, 10 or 14 layers (which corresponds to adding one block at a time to build ResNet-18) for the experiment. 
The performance of the overall network increases with the complexity of the saliency model but with diminishing returns. 

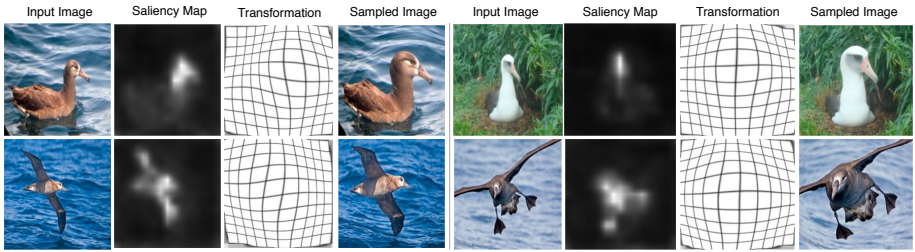
\myfigurew{figure_cub}{\textbf{Visualization of sampler behavior for the CUB-200 dataset}: 
We show the sampled images for a ResNet-50 trained with the saliency sampler in the CUB-200 dataset. The saliency amplifies relevant image regions such as the bird's head.}{1.0}


\subsection{CUB-200} 

To further prove that our model is useful across different datasets, we evaluated it in the CUB-200 dataset~\cite{wah2011caltech} (\refTbl{new_results}). Although the CUB-200 is also a fine-grained recognition dataset, it is significantly smaller and the images are better framed around subjects than in the iNaturalist dataset (see \refFig{figure_cub}). 

We used ResNet-50 as our task network and the initial $14$ layers of ResNet-18 as our saliency network. By adding our sampling layer we achieve a 2.9\% accuracy boost, which is less than the boost in iNaturalist, perhaps because objects of interest are more tightly cropped in CUB-200. Compared to DT-RAM \cite{li2017dynamic}, one of the top performing models in CUB-200, our approach outperforms the comparable $224\times224$ version of RN-50 DT-RAM by 1.7\%, using a simpler model. Our method is not as accurate as the $448\times448$ resolution version of DT-RAM, but the latter uses approximately 2 passes through a RN-50 on average and a larger input size leading to a higher computational cost.


\begin{table}[h!]
\centering
\begin{tabular}{l|c|c|c|c}
 & RN-50 & \textbf{RN-50+SS} & DT-RAM & DT-RAM \\ \toprule
\textbf{Res. (px)}  & 227 & \textbf{227} & 224 & 448 \\ 
\textbf{Top-1(\%)}  & 81.6 &  \textbf{84.5} & 82.8 & 86.0 \\ 
\end{tabular}
\caption{Performance improvements from the addition of our sampling layer on the CUB-200 dataset \cite{wah2011caltech}. \textbf{Res. (px)} refers to the input image resolution to the model.}
\label{tbl:new_results}
\end{table}

\section{Discussion} 
Adding our saliency sampler is most beneficial for image tasks where the important features are small and sparse, or appear across multiple image scales. The deformation introduced in the vicinity of the magnified regions could potentially discourage the network from strong deformations if another point of interest would be affected. This could be harmful for tasks such as text recognition. In practice, we observed that the learning process is able to deal with such situations well as it was capable of magnifying both collocated eyes without hindering the gaze prediction performance. That is particularly interesting as this task requires preservation of geometric information in the image.
The method proved to be easier to train than other approaches which modify spatial sampling, such as Spatial Transformer Networks~\cite{jaderberg2015spatial} or Deformable Convolutional Networks~\cite{dai2017deformable}.  These methods ofter performed closer to the baseline as they failed to find suitable parameters for their sampling strategy. 
The non-uniform approach to the magnification introduced by our saliency map also enables variability of zoom over the spatial domain. This together with the end-to-end optimization results in a performance benefit over uniformly magnified area-of-interest crops as observed in our fine-grained classification task. Unlike in the case of the iTracker~\cite{Khosla2016}, we do not require prior knowledge about the relevant image features in the task.

\section{Conclusion} 
We have presented the saliency sampler -- a novel layer for CNNs that can adapt the image sampling strategy to improve task performance while preserving memory allocation and computational efficiency for a given image processing task. We have shown our technique's effectiveness in locating and focusing on image features important for the tasks of gaze tracking and fine-grained object recognition. The method is simple to integrate into existing models and can be efficiently trained in an end-to-end fashion. Unlike some of the other image transformation techniques, our method is not restricted to a predefined number or size of important regions and it can redistribute sampling density across the entire image domain. At the same time, the parametrization of our technique by a single scalar attention map makes it robust against irrecoverable image degradation due to fold-overs or singularities. This leads to a superior performance in problems that require the recovery of small image features such as eyes or subtle differences between related animal species.
\\
\\
\textbf{Acknowledgment:} This research was funded by Toyota Research Institute. We acknowledge NVIDIA Corporation for hardware donations.

\bibliographystyle{splncs}
\bibliography{egbib}

\begin{thebibliography}{10}

\bibitem{krizhevsky2012imagenet}
Krizhevsky, A., Sutskever, I., Hinton, G.E.:
\newblock Imagenet classification with deep convolutional neural networks.
\newblock In: Conference on Neural Information Processing Systems. (2012)

\bibitem{iandola2016squeezenet}
Iandola, F.N., Han, S., Moskewicz, M.W., Ashraf, K., Dally, W.J., Keutzer, K.:
\newblock Squeezenet: Alexnet-level accuracy with 50x fewer parameters and< 0.5
  mb model size.
\newblock arXiv preprint arXiv:1602.07360 (2016)

\bibitem{simonyan2014very}
Simonyan, K., Zisserman, A.:
\newblock Very deep convolutional networks for large-scale image recognition.
\newblock arXiv preprint arXiv:1409.1556 (2014)

\bibitem{he2016deep}
He, K., Zhang, X., Ren, S., Sun, J.:
\newblock Deep residual learning for image recognition.
\newblock In: IEEE Conference on Computer Vision and Pattern Recognition.
  (2016)  770--778

\bibitem{ILSVRC15}
Russakovsky, O., Deng, J., Su, H., Krause, J., Satheesh, S., Ma, S., Huang, Z.,
  Karpathy, A., Khosla, A., Bernstein, M., Berg, A.C., Fei-Fei, L.:
\newblock {ImageNet Large Scale Visual Recognition Challenge}.
\newblock International Journal of Computer Vision (IJCV) \textbf{115}(3)
  (2015)  211--252

\bibitem{lowe2004distinctive}
Lowe, D.G.:
\newblock Distinctive image features from scale-invariant keypoints.
\newblock International Journal of Computer Vision (IJCV) \textbf{60}(2) (2004)
   91--110

\bibitem{ren2015faster}
Ren, S., He, K., Girshick, R., Sun, J.:
\newblock Faster r-cnn: Towards real-time object detection with region proposal
  networks.
\newblock In: Advances in Neural Information Processing System. (2015)  91--99

\bibitem{itti1998model}
Itti, L., Koch, C., Niebur, E.:
\newblock A model of saliency-based visual attention for rapid scene analysis.
\newblock IEEE Transactions on Pattern Analysis and Machine Intelligence
  \textbf{20}(11) (1998)  1254--1259

\bibitem{mnih2014recurrent}
Mnih, V., Heess, N., Graves, A.:
\newblock Recurrent models of visual attention.
\newblock In: Advances in Neural Information Processing System. (2014)
  2204--2212

\bibitem{ba2014multiple}
Ba, J., Mnih, V., Kavukcuoglu, K.:
\newblock Multiple object recognition with visual attention.
\newblock arXiv preprint arXiv:1412.7755 (2014)

\bibitem{eslami2016attend}
Eslami, S.A., Heess, N., Weber, T., Tassa, Y., Szepesvari, D., Hinton, G.E.,
  et~al.:
\newblock Attend, infer, repeat: Fast scene understanding with generative
  models.
\newblock In: Advances in Neural Information Processing Systems. (2016)
  3225--3233

\bibitem{fu2017look}
Fu, J., Zheng, H., Mei, T.:
\newblock Look closer to see better: {Recurrent} attention convolutional neural
  network for fine-grained image recognition.
\newblock In: Conf. on {Computer} {Vision} and {Pattern} {Recognition}. (2017)

\bibitem{jaderberg2015spatial}
Jaderberg, M., Simonyan, K., Zisserman, A.,  et~al.:
\newblock Spatial transformer networks.
\newblock In: Advances in Neural Information Processing System. (2015)
  2017--2025

\bibitem{dai2017deformable}
Dai, J., Qi, H., Xiong, Y., Li, Y., Zhang, G., Hu, H., Wei, Y.:
\newblock Deformable convolutional networks.
\newblock In: IEEE Conference on Computer Vision and Pattern Recognition.
  (2017)  764--773

\bibitem{zhou2016learning}
Zhou, B., Khosla, A., Lapedriza, A., Oliva, A., Torralba, A.:
\newblock Learning deep features for discriminative localization.
\newblock In: IEEE Conference on Computer Vision and Pattern Recognition, IEEE
  (2016)  2921--2929

\bibitem{li2017dense}
Li, J., Chen, Y., Cai, L., Davidson, I., Ji, S.:
\newblock Dense {Transformer} {Networks}.
\newblock arXiv:1705.08881 [cs, stat] (May 2017) arXiv: 1705.08881.

\bibitem{zheng2017learning}
Zheng, H., Fu, J., Mei, T., Luo, J.:
\newblock Learning multi-attention convolutional neural network for
  fine-grained image recognition.
\newblock In: IEEE International Conference on Computer Vision (ICCV). (2017)

\bibitem{xiao2015application}
Xiao, T., Xu, Y., Yang, K., Zhang, J., Peng, Y., Zhang, Z.:
\newblock The application of two-level attention models in deep convolutional
  neural network for fine-grained image classification.
\newblock In: IEEE Conference on Computer Vision and Pattern Recognition, IEEE
  (2015)  842--850

\bibitem{rosenfeld2016visual}
Rosenfeld, A., Ullman, S.:
\newblock Visual concept recognition and localization via iterative
  introspection.
\newblock In: Asian Conference on Computer Vision (ACCV), Springer (2016)
  264--279

\bibitem{selvaraju2017grad}
Selvaraju, R.R., Cogswell, M., Das, A., Vedantam, R., Parikh, D., Batra, D.:
\newblock Grad-cam: Visual explanations from deep networks via gradient-based
  localization.
\newblock In: IEEE Conference on Computer Vision and Pattern Recognition.
  (2017)  618--626

\bibitem{Khosla2016}
Khosla$^*$, A., Krafka$^*$, K., Kellnhofer, P., Kannan, H., Bhandarkar, S.,
  Matusik, W., Torralba, A.:
\newblock Eye tracking for everyone.
\newblock In: IEEE Conference on Computer Vision and Pattern Recognition, Las
  Vegas, USA (June 2016) $^*$ indicates equal contribution.

\bibitem{wang2017autoscaler}
Wang, S., Luo, L., Zhang, N., Li, J.:
\newblock {AutoScaler}: {Scale}-{Attention} {Networks} for {Visual}
  {Correspondence}.
\newblock In: British Machine Vision Conference ({BMVC}). (2017)

\bibitem{rubinstein2010comparative}
Rubinstein, M., Gutierrez, D., Sorkine, O., Shamir, A.:
\newblock A comparative study of image retargeting.
\newblock In: ACM Transactions on Graphics (TOG). Volume~29., ACM (2010)  160

\bibitem{wolf2007non}
Wolf, L., Guttmann, M., Cohen-Or, D.:
\newblock Non-homogeneous content-driven video-retargeting.
\newblock In: IEEE International Conference on Computer Vision (ICCV), IEEE
  (2007)  1--6

\bibitem{karni2009energy}
Karni, Z., Freedman, D., Gotsman, C.:
\newblock Energy-based image deformation.
\newblock In: Computer Graphics Forum. Volume~28., Wiley Online Library (2009)
  1257--1268

\bibitem{kaufmann2013finite}
Kaufmann, P., Wang, O., Sorkine-Hornung, A., Sorkine-Hornung, O., Smolic, A.,
  Gross, M.:
\newblock Finite element image warping.
\newblock In: Computer Graphics Forum. Volume~32., Wiley Online Library (2013)
  31--39

\bibitem{chen2010content}
Chen, R., Freedman, D., Karni, Z., Gotsman, C., Liu, L.:
\newblock Content-aware image resizing by quadratic programming.
\newblock In: Computer Vision and Pattern Recognition Workshops (CVPRW), IEEE
  (2010)  1--8

\bibitem{deng2009imagenet}
Deng, J., Dong, W., Socher, R., Li, L.J., Li, K., Fei-Fei, L.:
\newblock Imagenet: A large-scale hierarchical image database.
\newblock In: IEEE Conference on Computer Vision and Pattern Recognition, IEEE
  (2009)  248--255

\bibitem{van2017inaturalist}
Van~Horn, G., Mac~Aodha, O., Song, Y., Shepard, A., Adam, H., Perona, P.,
  Belongie, S.:
\newblock The inaturalist challenge 2017 dataset.
\newblock arXiv preprint arXiv:1707.06642 (2017)

\bibitem{zhou2015cnnlocalization}
Zhou, B., Khosla, A., A., L., Oliva, A., Torralba, A.:
\newblock {Learning Deep Features for Discriminative Localization.}
\newblock IEEE Conference on Computer Vision and Pattern Recognition (2016)

\bibitem{szegedy2016rethinking}
Szegedy, C., Vanhoucke, V., Ioffe, S., Shlens, J., Wojna, Z.:
\newblock Rethinking the inception architecture for computer vision.
\newblock In: Proceedings of the IEEE conference on computer vision and pattern
  recognition. (2016)  2818--2826

\bibitem{wah2011caltech}
Wah, C., Branson, S., Welinder, P., Perona, P., Belongie, S.:
\newblock The caltech-ucsd birds-200-2011 dataset.
\newblock (2011)

\bibitem{li2017dynamic}
Li, Z., Yang, Y., Liu, X., Zhou, F., Wen, S., Xu, W.:
\newblock Dynamic computational time for visual attention.
\newblock arXiv preprint arXiv:1703.10332 (2017)

\end{thebibliography}
\end{document}